\documentclass[10pt,twocolumn,letterpaper]{article}

\usepackage{cvm}
\usepackage{times}
\usepackage{epsfig}
\usepackage{graphicx}
\usepackage{amsmath}
\usepackage{amssymb}
\usepackage{subcaption} 
\usepackage{booktabs}
\usepackage{multirow}
\usepackage{float}
\usepackage{cite}
\usepackage{booktabs}
\usepackage{tabularx}
\usepackage{booktabs}
\usepackage{arydshln}
\usepackage{xcolor}
\usepackage{enumitem}  
\usepackage{arydshln}
\usepackage{graphicx}
\usepackage[pagebackref=true,breaklinks=true,letterpaper=true,colorlinks,bookmarks=false]{hyperref}

\newcommand{\keywords}[1]{{\bf \emph{Keywords: #1}}}

\cvmfinalcopy 

\def\cvmPaperID{301} 

\begin{document}

\title{BiCoR-Seg: Bidirectional Co-Refinement Framework for High-Resolution Remote Sensing Image Segmentation}

\author{Jinghao Shi\\
China University of Geosciences, Wuhan\\
Wuhan 430070, Hubei, China\\
{\small\url{shijinghao@cug.edu.cn}}
\and
Jianing Song\\
China University of Geosciences, Wuhan\\
Wuhan 430070, Hubei, China\\
{\small\url{songjn@cug.edu.cn}}
}

\maketitle

\begin{abstract}
High-resolution remote sensing image semantic segmentation (HRSS) is a fundamental yet critical task in the field of Earth observation. However, it has long faced the challenges of high inter-class similarity and large intra-class variability. Existing approaches often struggle to effectively inject abstract yet strongly discriminative semantic knowledge into pixel-level feature learning, leading to blurred boundaries and class confusion in complex scenes. To address these challenges, we propose Bidirectional Co-Refinement Framework for HRSS (BiCoR-Seg). Specifically, we design a Heatmap-driven Bidirectional Information Synergy Module (HBIS), which establishes a bidirectional information flow between feature maps and class embeddings by generating class-level heatmaps. Based on HBIS, we further introduce a hierarchical supervision strategy, where the interpretable heatmaps generated by each HBIS module are directly utilized as low-resolution segmentation predictions for supervision, thereby enhancing the discriminative capacity of shallow features. In addition, to further improve the discriminability of the embedding representations, we propose a cross-layer class embedding Fisher Discriminative Loss to enforce intra-class compactness and enlarge inter-class separability. Extensive experiments on the LoveDA, Vaihingen, and Potsdam datasets demonstrate that BiCoR-Seg achieves outstanding segmentation performance while offering stronger interpretability. The released code is available at \url{https://github.com/ShiJinghao566/BiCoR-Seg}.

\end{abstract}

\keywords{Remote Sensing, Image segmentation, Class embedding, Heatmap-driven, Bidirectional information synergy}

\section{Introduction}
High-resolution remote sensing image semantic segmentation (HRSS), as a fundamental task in the field of Earth observation~\cite{zhu2017deep}, holds significant application value in various real-world scenarios such as urban planning, land-use monitoring, ecological environment assessment, and disaster emergency response~\cite{luo2024rs,ma2019deep}. Its objective is to assign precise semantic labels to each pixel, thereby enabling the automatic identification of ground object categories and the fine depiction of spatial structures.

However, due to the diversity of ground object categories, significant variations in scale, complex textures, and drastic illumination changes in remote sensing images~\cite{luo2024rs,lei2023solo}, HRSS has long faced dual challenges of large intra-class variation and high inter-class similarity~\cite{sun2023semantic}. These characteristics make it difficult for models to learn feature representations that are highly discriminative and generalizable~\cite{wu2024semantic}, leading to common problems in complex scenes, such as blurred boundaries, category confusion, and fragmented regions. 

Researchers have conducted extensive explorations. Early convolutional neural network (CNN)-based methods~\cite{long2015fully,ronneberger2015u,lei2024ddranet} effectively extracted multi-scale local features through encoder–decoder architectures. However, due to the inherently limited receptive field of convolution operations, these methods struggled to capture global contextual information, resulting in insufficient capability to distinguish land-cover categories that depend on large-scale spatial layouts, such as roads and buildings. Subsequent works introduced dilated convolutions~\cite{yu2015multi}, attention mechanisms~\cite{bahdanau2014neural}, and Transformer architectures~\cite{vaswani2017attention} to enhance global perception by modeling long-range dependencies. Although these approaches improved feature representation capacity, they still relied on implicit semantic learning. Discriminative, category-level knowledge (e.g., "buildings" should exhibit rigid boundaries, "water bodies" should have smooth surfaces) remained buried within massive network parameters. Without an explicit mechanism to guide the alignment between pixel-level features and high-level semantics, models were prone to misclassification in regions with ambiguous features~\cite{wang2021exploring,lei2025yolov13realtimeobjectdetection}. 

Recent studies have attempted to incorporate class prototype mechanisms~\cite{zhang2022segvit,Nogueira_2024_WACV} to explicitly model semantic knowledge, enabling interactions between features and class representations within the semantic space. These methods have, to some extent, enhanced category separability and semantic consistency. However, such interactions are often unidirectional, as class embeddings influence feature aggregation during the decoding phase but cannot receive feedback from the pixel level for further refinement. Moreover, existing class embeddings are typically static and lack adaptability to specific image content, while their semantic spaces often lack explicit constraints. Therefore, establishing a tight and interpretable bidirectional optimization mechanism between semantic embeddings and visual features has become a key challenge for further improving HRSS performance.

To address the aforementioned issues, we propose the Bidirectional Co-Refinement Framework for HRSS (BiCoR-Seg). We posit that precise segmentation relies on establishing an explicit interaction mechanism between pixel-level features and category semantics during the forward propagation process of the network. Therefore, BiCoR-Seg abandons the traditional one-way decoding paradigm and introduces a heatmap-based bidirectional information synergy module (HBIS). In each interaction, HBIS first computes the similarity between the category embeddings and the feature map to generate class heatmaps. Subsequently, HBIS leverages these heatmaps to guide the category embeddings in perceiving the feature distributions, and updates their representations by selecting top-K high-response pixels. The enhanced category semantics then feed back into the feature maps by generating category-specific affine modulation parameters to adjust feature distributions. Furthermore, we design a dual supervision mechanism that employs a hierarchical heatmap loss to supervise the spatial localization accuracy and applies a Fisher discriminative loss to enforce intra-class compactness and inter-class separability of the category embedding space. This iterative interaction mechanism progressively enhances the discriminative power of features, enabling the network to simultaneously capture global semantic consistency and local boundary details, thereby significantly improving segmentation accuracy and generalization in complex scenes.

Overall, the contributions of this paper are as follows:
\begin{itemize}[noitemsep, topsep=0pt]

\item We propose the BiCoR-Seg framework, which achieves explicit semantic interaction and layer-wise collaborative refinement between feature and class embeddings, providing a novel and interpretable modeling paradigm for remote sensing image segmentation.
\item We design the HBIS module, which enables bidirectional information interaction driven by heatmaps to realize mutual guidance and semantic co-construction between features and class embeddings, thereby establishing an explicit and interpretable collaborative mechanism between the class embeddings and feature maps.
\item We introduce hierarchical heatmap deep supervision and a Fisher discriminant loss to enhance inter-layer semantic consistency and explicitly constrain the spatial structural distribution of class embeddings, significantly improving the discriminability of class embeddings and the separability of features.
\item Our BiCoR-Seg achieves highly competitive performance across three mainstream datasets, fully demonstrating the effectiveness and superiority of the proposed framework in complex scenarios.
\end{itemize}

\begin{figure*}[t]
    \centering
    \includegraphics[width=0.85\textwidth]{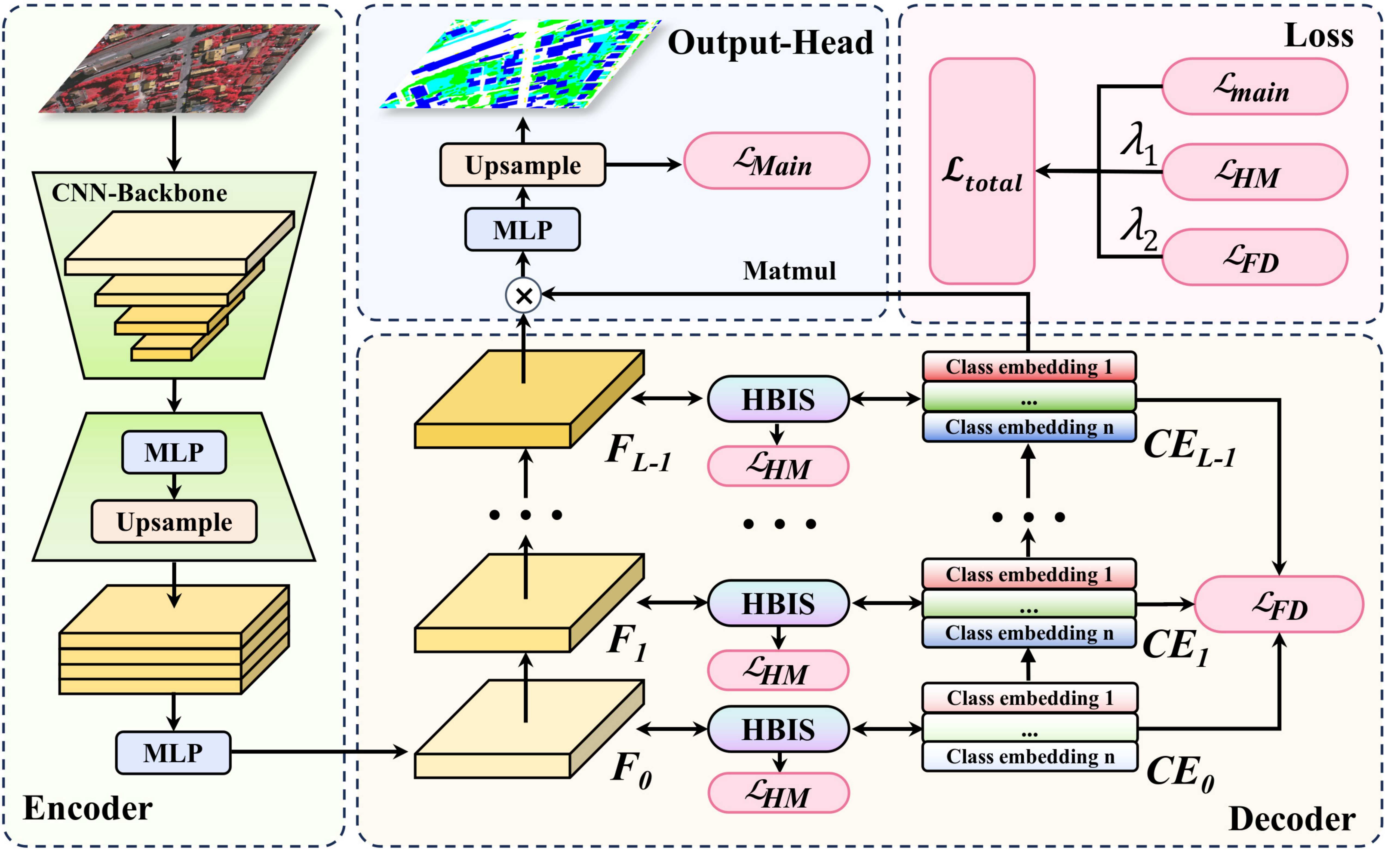}
    \caption{Overall architecture of the proposed BiCoR-Seg framework. 
    The proposed framework consists of an encoder for multi-scale feature extraction, a decoder composed of multiple cascaded HBIS modules, and an output head.}
    \label{fig:framework}
\end{figure*}

\section{Related Work}
\subsection{Deep Learning Based Remote Sensing Image Segmentation}
The introduction of deep learning has significantly advanced the development of semantic segmentation in remote sensing imagery. Early methods were primarily based on convolutional neural networks (CNNs). Representative of this stage, Fully Convolutional Networks (FCNs)~\cite{long2015fully} achieved pixel-level end-to-end prediction for the first time. Subsequent models such as U-Net~\cite{ronneberger2015u} and its variants, including UNet++~\cite{zhou2018unet++} and Attention U-Net~\cite{oktay2018attention}, effectively integrated high-level semantics with low-level spatial details through encoder–decoder architectures and skip connections.
To further enhance multi-scale perception, PSPNet~\cite{zhao2017pyramid} and the DeepLab series~\cite{chen2014semantic,chen2017deeplab,chen2019rethinking} introduced pyramid pooling and atrous (dilated) convolution structures. ConDSeg~\cite{lei2025condseg} improved boundary recognition through a contrastive-driven feature enhancement mechanism, while HRNet~\cite{sun2019deep} maintained high-resolution feature representations via parallel multi-resolution branches, demonstrating outstanding performance in fine-grained object segmentation.

To overcome the limitation of convolution’s local receptive field in global context modeling, researchers incorporated attention mechanisms~\cite{bahdanau2014neural} and Transformer~\cite{vaswani2017attention} architectures into segmentation tasks. Models such as SETR~\cite{zheng2021rethinking} and Segmenter~\cite{strudel2021segmenter} demonstrated the feasibility of pure Transformer-based designs; meanwhile, Swin Transformer~\cite{liu2021swin} and SegFormer~\cite{xie2021segformer} achieved a balance between performance and efficiency through hierarchical structures and lightweight decoders.
In the remote sensing domain, methods like AerialFormer~\cite{hanyu2024aerialformer} and UNetFormer~\cite{wang2022unetformer} combine the strengths of convolutional networks and Transformers, achieving a balance between local detail preservation and global semantic consistency. Hybrid architectures such as TransUNet~\cite{chentransunet}, LeViT-UNet~\cite{xu2023levit}, and Swin-UNet~\cite{cao2022swin} further strengthen cross-level feature interaction, exhibiting superior performance in multi-scale object recognition and complex scene segmentation. Collectively, these studies constitute the technological foundation of modern remote sensing segmentation models.

\subsection{Explicit Semantic Modeling Methods}
Recent studies have begun exploring explicitly semantic-guided paradigms to more directly integrate category-level semantic knowledge into the segmentation process. The core idea originates from the query-based object detector DETR~\cite{zhu2020deformable}, which introduces a set of learnable class prototypes or class queries. These methods maintain a dedicated embedding vector for each semantic category, serving as a carrier of high-level semantic concepts that interact with pixel-level features during the network's decoding stage. In the field of general segmentation, MaskFormer~\cite{cheng2021per} and its subsequent improvement Mask2Former~\cite{cheng2022masked} have achieved remarkable success by reformulating the segmentation task as a query-based mask classification problem. K-Net~\cite{zhang2021k} further extends this idea by employing a set of dynamically updated kernels for instance segmentation, while SegViT~\cite{zhang2022segvit} utilizes learnable class queries to compute attention maps for generating explicit semantic masks. In the remote sensing domain, the Prototypical Contrastive Network (PCN)~\cite{Nogueira_2024_WACV} enhances intra-class compactness in binary segmentation tasks through prototype contrastive constraints, whereas CenterSeg~\cite{zhang2025center} further designs a classifier based on category centers and multiple prototypes.

However, in existing methods, the interaction between semantics and visual features is typically unidirectional and shallow, preventing high-level semantic knowledge from effectively guiding low-level feature learning. Moreover, most prior class embedding approaches are static, meaning that the class prototypes remain unchanged during inference, independent of the image context, and lack hierarchical adaptability or explicit regularization within the embedding space.
In contrast to the above works, our proposed BiCoR-Seg framework establishes a bidirectional and iterative information flow across all decoder stages via the HBIS module, leveraging class heatmaps to ensure that semantic knowledge deeply and continuously guides the entire feature learning process. Furthermore, we introduce hierarchical class heatmap supervision and a Fisher discriminant loss, which jointly impose direct and effective constraints on the cooperative refinement between features and embeddings from both spatial localization and semantic distribution perspectives.

\begin{figure*}[t]
    \centering
    \includegraphics[width=\textwidth]{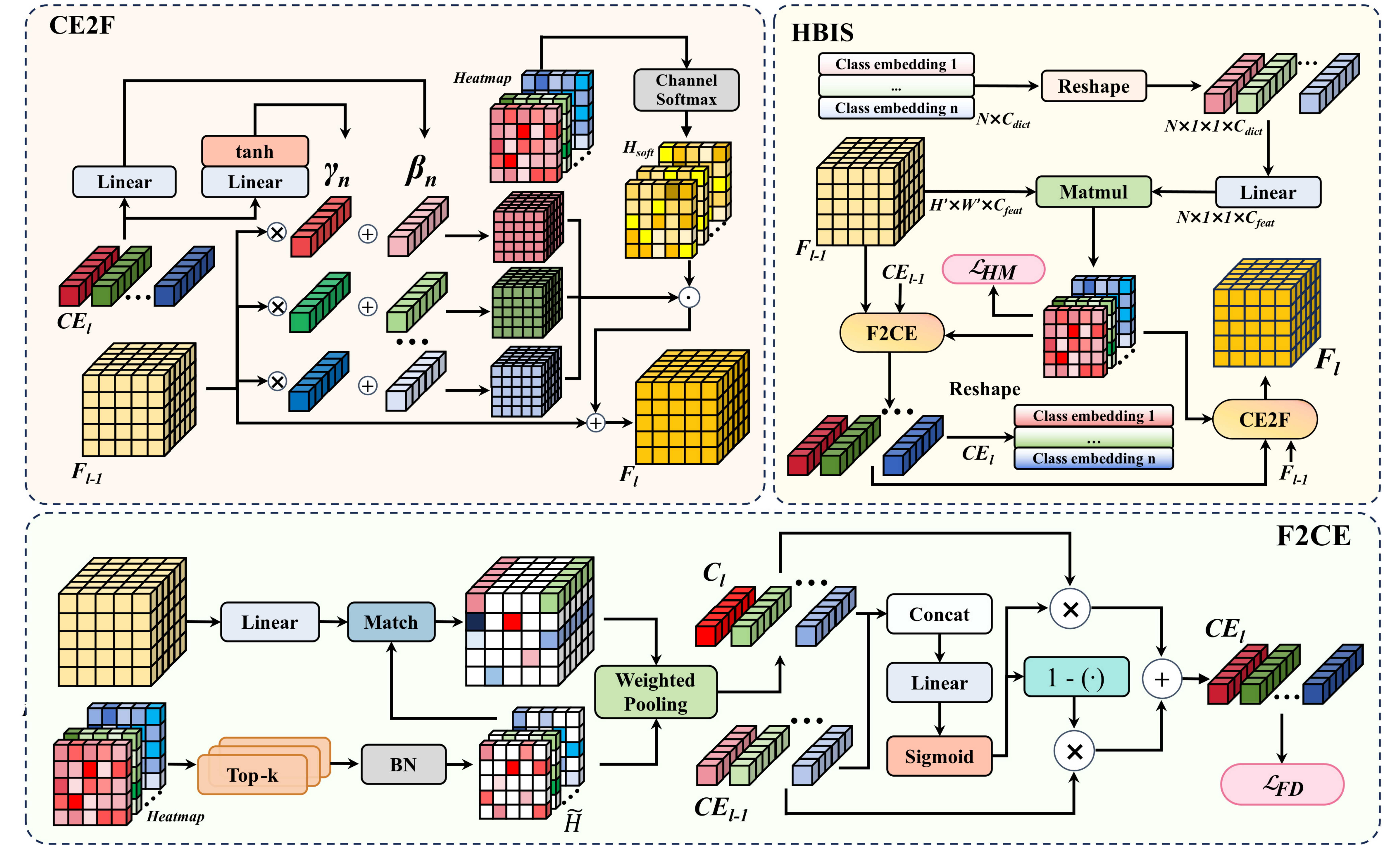}
    \caption{
        Architecture of the proposed HBIS module. 
        The HBIS module is composed of two information pathways: 
        Feature-to-Class Embedding (F2CE) and Class Embedding-to-Feature (CE2F). 
        Three types of heatmaps are utilized in the bidirectional interaction: 
        the raw class heatmap, the normalized heatmap $\tilde{H}$ (with a BN layer for normalization),  and the Softmax-normalized heatmap $H^{soft}$ for semantic fusion.
    }
    \label{fig:hbis}
\end{figure*}

\section{Method}

\subsection{Overall Architecture}
To address the issues of large intra-class variance and small inter-class differences in high-resolution remote sensing image semantic segmentation (HRSS), which often result in recognition ambiguity and blurred boundaries, and to enable interactive fusion between class embeddings and pixel-level features during the decoding stage, we propose a Bidirectional Cooperative Refinement framework for HRSS(BiCoR-Seg). This framework progressively injects discriminative semantic concepts into the pixel feature learning process, thereby achieving deep integration between semantic embeddings and visual representations.

As illustrated in Fig.~\ref{fig:framework}, the proposed framework consists of three main components: an encoder for extracting multi-scale visual features, an iterative decoder that performs cooperative refinement between features and semantics, and an output head that produces the final segmentation results. Specifically, given an input image $I \in \mathbb{R}^{H \times W \times 3}$, the encoder first extracts multi-scale features and fuses them through an aggregation module composed of MLPs and upsampling operations, resulting in a high-resolution feature map $F_0$ and an initial class embedding $CE_0$.
Subsequently, $F_0$ and $CE_0$ are refined through an $L$-layer decoder constructed with HBIS modules, which perform bidirectional cooperative refinement between pixel features and class embeddings. Let $CE_l = \{CE_{l,n}\}_{n=1}^{N}$ denote the set of class embeddings output by the $l$-th decoder layer, where $CE_{l,n}$ represents the embedding vector of the $n$-th class at layer $l$, and $N$ is the total number of semantic classes.
Finally, the output head performs matrix multiplication between the final feature map $F_L$ and the class embeddings $CE_L$, and the result is upsampled to obtain the final pixel-level probability prediction map $P$.

In the following sections, we focus on the core component of the framework, the Heatmap-driven Bidirectional Information Synergy module (HBIS), and provide a detailed description of the hierarchical heatmap supervision loss and Fisher discriminant loss that are specifically designed to optimize this architecture.

\subsection{Heatmap-driven Bidirectional Information Synergy Module (HBIS)}
Although current explicit semantic modeling methods introduce class-prototype embeddings, their interactions are often unidirectional: the class embeddings only participate in feature aggregation at the end of decoding, lacking positive guidance for shallow feature learning. Moreover, the updating process of class embeddings is static and independent of the image content, preventing high-level semantic knowledge from actively influencing the entire feature learning process.

To address this issue, we design HBIS, as illustrated in Fig.~\ref{fig:hbis}, to establishes an interpretable and effective bidirectional communication between class embeddings and pixel-level features. In the following subsection, we first introduce the core component, namely the generation of class heatmaps, and then, based on this foundation, elaborate on two enhancement pathways within HBIS: the feature-to-class embedding update path (F2CE) and the class embedding-to-feature enhancement path (CE2F).
\paragraph{Class Heatmap Generation.}
To initiate bidirectional communication, it is first necessary to establish a correspondence between abstract class concepts and their spatial locations in the image. Therefore, we design a class heatmap that projects each class embedding into the image space, generating a spatial confidence map that reveals the potential distribution of each class across the image.

We dynamically activate the functionality of the class embeddings, enabling them to serve as query vectors. Specifically, the class embeddings from the previous layer, denoted as $CE_{l-1}$, are transformed through a learnable linear layer $\text{Linear}(\cdot)$ into a set of vectors whose dimensions match the channel dimension of the feature map $C_{\text{feat}}$. By computing the dot-product similarity between each pixel feature and all class embeddings, we obtain the response intensity of the $n$-th class at the spatial location $(x, y)$:
\begin{equation}
H_{l,n}(x, y) = \sigma \big( F_{l-1}(x, y) \cdot \text{Linear}(CE_{l-1,n})^{\top} \big),
\end{equation}
where $\sigma(\cdot)$ denotes the Sigmoid activation function. The output value $H_{l,n}(x, y) \in [0,1]$ represents the confidence that the pixel belongs to the $n$-th class. Consequently, a class heatmap for the current layer is formed as $H_l \in \mathbb{R}^{H' \times W' \times N}$, whose resolution is consistent with that of the feature map and is utilized for subsequent bidirectional information collaboration.

\paragraph{Feature Map to Class Embedding Updating (F2CE).}

The core objective of the F2CE pathway is to endow class embeddings with image adaptivity, thereby progressively refining generic semantic priors into contextual representations tightly related to the content of the current image. At layer $l$, this pathway aggregates, for each category, the most relevant visual features from the feature map $F_{l-1}$. The input class embedding $CE_{l-1}$ thus incorporates the image-specific contextual information and is updated into a more discriminative $CE_{l-1}$.

This procedure consists of two steps. First is region-guided feature pooling: for each class, we use its heatmap $H_{l,n}$ to select the Top-$K$ high-response pixels and aggregate them into a region $\Omega_{l,n}$. Using the normalized heatmap values as weights, the pixel features within these regions are projected by a linear layer, $\mathrm{Linear}(\cdot)$, into the class-embedding dimensionality $C_{\text{class}}$, followed by weighted pooling. This ensures that the information aggregation concentrates on the most relevant visual cues while suppressing the interference from irrelevant areas. Finally, we obtain the contextual feature vector for class $n$ at the current layer, $C_{l,n}$, which aggregates the most representative pixel information of class $n$ in the current image, as formulated below
\begin{equation}
\tilde{H}_{l,n}(x,y)=
\begin{cases}
\dfrac{H_{l,n}(x,y)}{\sum\limits_{(u,v)\in\Omega_n}H_{l,n}(u,v)+\varepsilon}, & (x,y)\in\Omega_n,\\[10pt]
0, & (x,y)\notin\Omega_n,
\end{cases}
\label{eq:heatmap_norm}
\end{equation}
\begin{equation}
C_{l,n}=\sum\limits_{(x,y)\in\Omega_n}\tilde{H}_{l,n}(x,y)\cdot \text{Linear}(F_{l-1}(x,y)).
\label{eq:context_vector}
\end{equation}
Here, $\tilde{H}_{l,n}(x,y)$ denotes the normalized class heatmap value of the $n$-th class at pixel position $(x,y)$ in the $l$-th layer.

Directly adding or replacing the contextual vector $C_l$ with the class embedding $CE_{l-1}$ may lead to instability during training. Therefore, we introduce a gated update mechanism to achieve stable and adaptive fusion. Specifically, we compute an adaptive gating vector $G_l$ by concatenating the previous class embedding and the current contextual feature vector, and then passing the result through a learnable linear layer $\text{Linear}(\cdot)$ to dynamically determine the ratio between information preservation and absorption:
\begin{equation}
G_l = \sigma\big(\text{Linear}([CE_{l-1} \Vert C_l])\big),
\label{eq:gating}
\end{equation}
where $\Vert$ denotes vector concatenation, and $G_l \in \mathbb{R}^N$. Using this gating vector $G_l$, we perform weighted fusion between historical semantic information and current contextual information to generate updated class embeddings. For the $n$-th class embedding at the $l$-th layer, the update rule is defined as:
\begin{equation}
CE_{l,n} = (1 - G_{l,n}) \cdot CE_{l-1,n} + G_{l,n} \cdot C_{l,n}.
\end{equation}
Here, $G_{l,n}$ represents the gating coefficient of $G_l$ for the $n$-th class. This mechanism enables each class embedding to maintain semantic stability while adapting dynamically to the specific image content.

\paragraph{Class Embedding to Feature Map Updating (CE2F).}

When the class embeddings are updated through the F2CE pathway, they carry more accurate and context-aware semantic knowledge. The goal of the CE2F pathway is to write the semantic knowledge learned by $CE_l$ back to the pixel-level feature maps, guiding and enhancing those ambiguous features to become more category-discriminative.

This process also consists of two steps. First is the class-specific feature modulation. We assume that objects of different classes should have distinct feature distributions. Therefore, using the updated class embeddings $CE_l$, we learn a pair of affine transformation parameters for each class $CE_{l,n}$, namely the channel-wise scaling factor $\gamma_n$ and bias $\beta_n$. These affine parameters are generated as follows:
\begin{equation}
\gamma_n = 1 + \tanh\!\big(\mathrm{Linear}(CE_{l,n})\big),
\label{eq:gamma}
\end{equation}
\begin{equation}
\beta_n = \mathrm{Linear}(CE_{l,n}),
\label{eq:beta}
\end{equation}
where $\tanh(\cdot)$ denotes the hyperbolic tangent activation function. Both $\gamma_n$ and $\beta_n$ belong to $\mathbb{R}^{C_{\text{feat}}}$.These parameters can be viewed as the encapsulation of the $n$-th semantic concept in the feature space. The original features are modulated by these class-specific affine parameters to form the modulated features $\tilde{F}_{l,n}$. For each pixel $(x,y)$, the updated feature vector $\tilde{F}_{l,n}(x,y)$ is computed as:
\begin{equation}
\tilde{F}_{l,n}(x,y) = \gamma_n \odot F_{l-1}(x,y) + \beta_n,
\label{eq:feature_modulation}
\end{equation}
where $F_{l-1}(x,y) \in \mathbb{R}^{C_{\text{feat}}}$ represents the feature vector at position $(x,y)$, and $\odot$ denotes channel-wise multiplication.
Next comes the heatmap-guided modulation feature fusion. The degree to which each pixel should be influenced by a specific class-modulated feature is controlled by the class heatmap. We apply a Softmax normalization along the class dimension of the heatmap $H_l$ to obtain the per-pixel class assignment probability $H^{\text{soft}}_l$:
\begin{equation}
H^{\mathrm{soft}}_{l,n}(x,y) = 
\frac{\exp\!\big(H_{l,n}(x,y)\big)}
{\sum\limits_{k=1}^{N} \exp\!\big(H_{l,k}(x,y)\big)},
\label{eq:heatmap_softmax}
\end{equation}
In this context, $H^{\mathrm{soft}}_{l,n}(x,y)$ denotes the Softmax-normalized heatmap value of the $n$-th class at pixel position $(x,y)$ in the $l$-th layer, representing the probability that this pixel belongs to class $n$. Then, we use these probabilities as weights to fuse the modulated features from all classes and add them back to the original features to form the enhanced representation:
\begin{equation}
F_l(x,y) = \alpha F_{l-1}(x,y) 
+ (1 - \alpha) 
\sum_{n=1}^{N} H^{\text{soft}}_{l,n}(x,y) \cdot \tilde{F}_{l,n}(x,y),
\label{eq:final_fusion}
\end{equation}
where $\alpha$ is a learnable parameter, and $H^{\text{soft}}_{l,n}(x,y)$ denotes the probability that the pixel at position $(x,y)$ belongs to the $n$-th class. Through the CE2F pathway, each pixel-level feature representation is directly guided by the most probable class-specific semantic knowledge, effectively mitigating issues of inter-class similarity and intra-class ambiguity.
\begin{table*}[h]
\centering
\caption{Comparison with other methods on the LoveDA dataset. "Agri." denotes Agriculture.}
\label{tab:loveda}
\renewcommand{\arraystretch}{1.3}
\setlength{\tabcolsep}{5pt}
\resizebox{\textwidth}{!}{%
\begin{tabular}{lcccccccccc}
\toprule
\textbf{Method} & \textbf{Venue} & \textbf{Background.} & \textbf{Building} & \textbf{Road} & \textbf{Water} & \textbf{Barren} & \textbf{Forest} & \textbf{Agri.} & \textbf{mIoU} \\
\midrule
FCN~\cite{Long_2015_CVPR} & CVPR 2015 & 42.6 & 49.5 & 48.1 & 73.1 & 11.8 & 43.5 & 58.3 & 46.7 \\
U\mbox{-}Net~\cite{ronneberger2015u} & MICCAI 2015 & 43.1 & 52.7 & 52.8 & 73.1 & 10.3 & 43.0 & 59.9 & 47.8 \\
PSPNet~\cite{zhao2017pyramid} & ISPRS 2016 & 44.4 & 52.1 & 53.5 & 76.5 & 9.7 & 44.1 & 57.9 & 48.3 \\
LinkNet~\cite{chaurasia2017linknet} & CVPR 2017 & 43.6 & 52.1 & 52.5 & 76.9 & 12.2 & 45.1 & 57.3 & 48.5 \\
UNet++~\cite{zhou2018unet++} &MICCAI  2018& 42.9 & 52.6 & 52.8 & 74.5 & 11.4 & 44.4 & 58.8 & 48.2 \\
DeepLabv3+~\cite{chen2018encoder} & ECCV 2018 & 43.0 & 50.9 & 52.0 & 74.4 & 10.4 & 44.2 & 58.5 & 47.6 \\
SemanticFPN~\cite{Kirillov_2019_CVPR} & CVPR 2019 & 42.0 & 51.5 & 53.4 & 74.7 & 11.2 & 44.6 & 58.7 & 48.2 \\
FarSeg~\cite{zheng2020foreground} & CVPR 2020 & 43.4 & 51.8 & 53.3 & 76.1 & 10.8 & 43.2 & 58.6 & 48.2 \\ 
SegFormer~\cite{xie2021segformer} & NeurIPS 2021 & 42.2 & 56.4 & 50.7 & 78.5 & 17.2 & 45.2 & 53.8 & 49.1 \\ 
BANet~\cite{wang2022novel} & GRSL 2022 & 43.7 & 51.5 & 51.1 & 76.9 & 16.6 & 44.9 & 62.5 & 49.6 \\
UNetFormer~\cite{wang2022unetformer} & ISPRS 2022 & 44.7 & 58.8 & 54.9 & 79.6 & 20.1 & 46.0 & 62.5 & 52.4 \\
RSSFormer~\cite{xu2023rssformer} &  TIP 2023 & \textbf{52.4} & 60.7 & 55.2 & 76.3 & 18.7 & 45.4 & 58.3 & 52.4 \\ 
TransUNet~\cite{chentransunet} & MIA 2024 & 43.0 & 56.1 & 53.7 & 78.0 & 9.3 & 44.9 & 56.9 & 48.9 \\
Hi-ResNet~\cite{chen2024hi}& JSTARS 2024 & 46.7 & 58.3 & 55.9 & 80.1 & 17.0 & 46.7 & 62.7 & 52.5 \\
SFA-Net~\cite{hwang2024sfa} & RS 2024 & 48.4 & 60.3 & 59.1 & 81.9 & 24.1 & 46.2 & 64.0 & 54.9 \\
AFENet~\cite{gao2025adaptive} & TGRS 2025 & 47.4 & 59.2 & \textbf{59.2} & 81.6 & 21.4 & 48.7 & 66.4 & 54.8 \\
DMA-Net~\cite{deng2025dma} & RS 2025 & 49.3 & 60.1 & 58.7 & \textbf{82.3} & 17.3 & 47.1 & 64.8 & 54.2 \\
\midrule
\textbf{Ours} & -- & 48.1 & \textbf{60.9} & 58.7 & 80.8 & \textbf{24.4} & \textbf{48.7} & \textbf{66.8} & \textbf{55.5} \\
\bottomrule
\end{tabular}
} 
\end{table*}

For the hierarchical heatmap supervision loss $\mathcal{L}_{HM}$, 
the effectiveness of each HBIS module highly depends on the quality of the generated class heatmaps.
To ensure that heatmaps at shallow layers can learn meaningful semantic localization ability,
we regard them as low-resolution segmentation predictions and apply deep supervision.
At each decoder layer $l$, the heatmap $H_l$ is upsampled to the ground-truth size
and constrained with standard segmentation losses:
\begin{equation}
\mathcal{L}_{HM} = 
\sum_{l=1}^{L} \mathcal{L}_{CE}(\text{Up}(H_l), Y)
+ \mathcal{L}_{Dice}(\text{Up}(H_l), Y)
\label{eq:hm_loss},
\end{equation}
where $\text{Up}(\cdot)$ denotes the upsampling operation that resizes the prediction map to the same spatial resolution as the ground truth $Y$. This deep supervision strategy effectively alleviates the gradient vanishing problem 
and strengthens the model's capability of learning discriminative spatial features 
in shallow layers.

\subsection{Multi-level Cooperative Supervision}
To ensure that BiCoR-Seg can learn highly discriminative features and semantic representations, 
we design a composite multi-level loss function. The total loss is defined as:
\begin{equation}
\mathcal{L}_{total} = \mathcal{L}_{main} + \lambda_1 \mathcal{L}_{HM} + \lambda_2 \mathcal{L}_{FD}
\label{eq:total_loss},
\end{equation}
\begin{equation}
\mathcal{L}_{main} = \mathcal{L}_{CE}(P, Y) + \mathcal{L}_{Dice}(P, Y).
\label{eq:main_loss}
\end{equation}
Although HBIS achieves semantic alignment at the spatial level, 
class embeddings may still overlap in semantic space,
that is, embeddings of different classes become too close, 
leading to ambiguous class discrimination. 
To address this, we introduce a Fisher Discriminant Constraint $\mathcal{L}_{FD}$,
which explicitly enhances inter-class separability and intra-class compactness 
from a geometric perspective of the feature space. 
For the class embedding set at the $l$-th layer, 
$CE_l = \{ CE_{l,n} \}_{n=1}^{N}$, 
we compute the within-class scatter $S_w^{(l)}$ 
and between-class scatter $S_b^{(l)}$ within a mini-batch as follows:
\begin{equation}
S_w^{(l)} = 
\frac{1}{BN} \sum_{n=1}^{N} \sum_{b=1}^{B}
\| CE_{l,n}^{(b)} - \mu_n^{(l)} \|_2^2,
\label{eq:sw}
\end{equation}
where $\mu_n^{(l)}$ and $\mu^{(l)}$ denote the class center and the overall mean
at the $l$-th layer, respectively:
\begin{equation}
\mu_n^{(l)} = \frac{1}{B} \sum_{b=1}^{B} CE_{l,n}^{(b)},
\label{eq:mu_n}
\end{equation}
\begin{equation}
\mu^{(l)} = \frac{1}{N} \sum_{n=1}^{N} \mu_n^{(l)}.
\label{eq:mu}
\end{equation}

The Fisher Discriminant loss $\mathcal{L}_{FD}$ aims to minimize the ratio between 
the within-class scatter and the between-class scatter, 
encouraging embeddings of the same category to be more compact 
and those of different categories to be farther apart in the feature space. 
It is formulated as:
\begin{equation}
\mathcal{L}_{FD}^{(l)} = 
\frac{S_w^{(l)}}{S_b^{(l)} + \epsilon}.
\label{eq:lfd_layer}
\end{equation}
We apply this loss to the output $CE_l$ of each decoder layer 
and compute a weighted summation to ensure that 
a highly discriminative semantic space is maintained 
from shallow to deep layers:
\begin{equation}
\mathcal{L}_{FD} = 
\sum_{l=1}^{L} \mathcal{L}_{FD}^{(l)}.
\label{eq:lfd_total}
\end{equation}

Through the cooperative effect of the hierarchical heatmap supervision loss 
$\mathcal{L}_{HM}$ and the Fisher Discriminant constraint 
$\mathcal{L}_{FD}$, 
BiCoR-Seg achieves a comprehensive optimization objective, 
which significantly enhances segmentation accuracy 
and class discrimination capability, 
while exhibiting superior generalization stability in complex scenarios.
\begin{table*}[t]
\centering
\caption{Comparison with other methods on the ISPRS Vaihingen dataset. "Imp. surf." and "Low veg." stand for Impervious surfaces and Low vegetation, respectively.}
\label{tab:vaihingen}
\renewcommand{\arraystretch}{1.3}
\setlength{\tabcolsep}{6pt} 
\begin{tabular*}{\textwidth}{@{\extracolsep{\fill}}lccccccccc}
\toprule
\textbf{Method} & \textbf{Venue} & \textbf{Imp. surf.} & \textbf{Building} & \textbf{Low veg.} & \textbf{Tree} & \textbf{Car} & \textbf{mIoU} & \textbf{OA} & \textbf{mF1} \\
\midrule
U-Net~\cite{ronneberger2015u} & MICCAI 2015 & 84.33 & 86.48 & 73.13 & 83.89 & 40.82 & 60.92 & 82.02 & 73.73 \\
DANet~\cite{fu2019dual} & CVPR 2019 & 91.13 & 94.82 & 83.47 & 88.92 & 62.98 & 74.52 & 89.52 & 84.27 \\
TreeUNet~\cite{yue2019treeunet} & ISPRS 2019 & 92.50 & 94.90 & 83.60 & 89.60 & 85.90 & 80.92 &90.40& 89.30\\
SegViT~\cite{zhang2022segvit} & NeurIPS 2022 & 91.97 & 95.26 & 82.24 & 90.84 & 80.68 & 79.35 & 90.50 & 88.20 \\
CTFNet~\cite{wu2023ctfnet} & GRSL 2023 & 90.69 & 94.35 & 81.66 & 87.27 & 82.72 & 77.84 & 88.60 & 87.34 \\
SFA-Net~\cite{hwang2024sfa} & RS 2024 & 93.50 & 96.30 & 85.40 & 90.20 & \textbf{90.70} & 84.06 & - & 91.20 \\
LSRFormer~\cite{zhang2024lsrformer} & TGRS 2024 & 93.84 & \textbf{96.40} & \textbf{85.82} & 90.70 & 90.79 & 84.56 & 91.90 & 91.50 \\
AFENet~\cite{gao2025adaptive} & TGRS 2025 & 96.90 & 95.72 & 85.07 & 90.64 & 89.37 & 84.55 & 91.67 & 91.54 \\
\midrule
\textbf{Ours} & -- & \textbf{97.36} & 96.11 & 85.78 & \textbf{91.08} & 89.40 & \textbf{85.38} & \textbf{92.10} & \textbf{91.94} \\
\bottomrule
\end{tabular*}
\end{table*}

\begin{table*}[t]
\centering
\caption{Comparison with other methods on the ISPRS Potsdam dataset.}
\label{tab:potsdam}
\renewcommand{\arraystretch}{1.3}
\setlength{\tabcolsep}{6pt}
\begin{tabular*}{\textwidth}{@{\extracolsep{\fill}}lccccccccc}
\toprule
\textbf{Method} & \textbf{Venue} & \textbf{Imp. surf.} & \textbf{Building} & \textbf{Low veg.} & \textbf{Tree} & \textbf{Car} & \textbf{mIoU} & \textbf{OA} & \textbf{mF1} \\
\midrule
U-Net~\cite{ronneberger2015u} & MICCAI 2015 & 86.54 & 88.73 & 72.09 & 79.49 & 46.67 & 60.46 & 85.29 & 75.64 \\
DANet~\cite{fu2019dual} & CVPR 2019 & 91.00 & 95.60 & 86.10 & 87.60 & 84.30 & 80.30 & 89.10 & 88.90 \\
TreeUNet~\cite{yue2019treeunet} & ISPRS 2019 & 93.10 & 97.30 & 86.80 & 87.10 & 95.80 & 85.50 & 90.70 & 92.00\\
SegViT~\cite{zhang2022segvit} & NeurIPS 2022 & 92.45 & 97.01 & 89.77 & 89.98 & 95.23 & 86.85 & 91.20 & 92.89 \\
CTFNet~\cite{wu2023ctfnet} & GRSL 2023 & 91.48 & 96.30 & 86.04 & 87.23 & 92.48 & 83.20 & 89.38 & 90.70 \\
LSRFormer~\cite{zhang2024lsrformer} & TGRS 2024 & 94.08 & 97.33 & 88.63 & 89.82 & 97.05 & 87.80 & 91.90 & 93.40 \\
SFA-Net~\cite{hwang2024sfa} & RS 2024 & \textbf{95.00} & \textbf{97.50} & 88.30 & 89.60 & 97.10 & 87.60 & - & 93.50 \\
AFENet\cite{gao2025adaptive} & TGRS 2025 & 94.57 & 96.86 & 88.17 & 89.78 & 96.83 & 87.50 & 92.00 & 93.24 \\
\midrule
\textbf{Ours} & -- & 94.82 & 97.38 & \textbf{91.83} & \textbf{91.40} & \textbf{97.52} & \textbf{89.87} & \textbf{92.10} & \textbf{94.59} \\
\bottomrule
\end{tabular*}
\end{table*}
\section{Experiments}

\subsection{Datasets and Evaluation Metrics}

We conduct extensive experiments and evaluations on three challenging high-resolution remote sensing image segmentation benchmark datasets. 
To ensure reproducibility and fair comparison, all datasets are divided following the standard protocols provided by the open-source remote sensing segmentation framework GeoSeg~\cite{DBLP:journals/corr/abs-1809-03175}.

\textbf{LoveDA (Land-cover Domain Adaptive)~\cite{wang2021loveda}.} A large-scale dataset containing both urban and rural scenes, with a spatial resolution of 0.3\,m and covering seven land-cover categories. 
This dataset exhibits diverse scenes and severe class imbalance, posing high demands on the model's domain adaptation capability and robustness. 
It contains 2522 images for training, 1669 images for validation, and 1796 images for testing. 
Since the ground-truth labels of the test set are not publicly available, we follow the official evaluation protocol and submit our predictions to the online evaluation server to obtain the final results.

\textbf{ISPRS Vaihingen.~\cite{rottensteiner2012isprs}} 
This dataset consists of 33 high-resolution aerial images with a ground sampling distance (GSD) of 9\,cm. 
It contains six classes in total: five foreground categories (impervious surfaces, buildings, low vegetation, trees, and cars) and one background class. 
Following the GeoSeg split protocol, 15 images are used for training, one image (ID 30) for validation, and 17 images (IDs 2, 4, 6, etc.) for testing~\cite{wang2022unetformer}. 
All images are cropped into non-overlapping patches of $1024 \times 1024$ pixels for training and inference.

\textbf{ISPRS Potsdam.~\cite{rottensteiner2012isprs}} 
This dataset includes 38 aerial images with even higher resolution, each with a size of $6000 \times 6000$ pixels. 
The category definitions are consistent with those of the Vaihingen dataset. 
Following the official split, after removing one mislabeled image (ID 7\_10), we use 22 images for training, one image (ID 2\_10) for validation, and 14 images (IDs 2\_13, 2\_14, 3\_13, etc.) for testing~\cite{wang2022unetformer}. 
All images are also cropped into patches of $1024 \times 1024$ pixels before being fed into the model.

To quantitatively evaluate the segmentation performance, we adopt three widely used metrics in semantic segmentation tasks, following previous studies~\cite{everingham2007pascal,cordts2016cityscapes,long2015fully}: mean Intersection-over-Union (mIoU), Overall Accuracy (OA), and mean F1-Score (mF1).

\subsection{Implementation Details}
All experiments are conducted on an NVIDIA RTX 4090 GPU under the PyTorch framework. 
We adopt ConvNeXt-B~\cite{liu2022convnet} as the backbone network and initialize it with weights pre-trained on ImageNet. 
Two HBIS layers are inserted into the decoder stage; within each layer, a Top-$K$ strategy is used to select the most discriminative regions for semantic aggregation, with the parameter set to $0.02$, i.e., the top $2\%$ high-confidence pixels per class are retained during training. 
We use the Adam optimizer~\cite{kingma2014adam} with an initial learning rate of $0.8\times10^{-4}$ and apply a cosine annealing schedule to dynamically adjust the learning rate. 
The batch size is set to $8$. 
The weights of the composite loss are empirically set to $\lambda_{1}=0.1$ and $\lambda_{2}=0.1$.

\subsection{Comparison with Other Methods}
On the most challenging LoveDA dataset, our model achieves an mIoU of 55.5\%, reaching a new SOTA level, as shown in Table~\ref{tab:loveda}.
The model performs particularly well on the ``Agriculture'' and ``Forest'' categories, with IoUs of 66.8\% and 48.7\%, respectively, indicating that our approach effectively handles large-scale and texture-complex natural scenes.

On the ISPRS Vaihingen dataset, our model achieves an mF1 score of 91.94\%, as shown in Table~\ref{tab:vaihingen}. The advantage is particularly evident for the ``Imp.\ surf'' and ``Tree'' classes, where the F1 scores reach 97.36\% and 91.08\%, respectively, fully demonstrating that BiCoR-Seg excels at modeling fine structures and complex boundaries in high-resolution aerial imagery.

On the ISPRS Potsdam dataset, our model also delivers strong results, with an mF1 of 94.59\%, as shown in Table~\ref{tab:potsdam}.
It achieves notable improvements in distinguishing the highly confusable "Low veg." and "Tree" categories. 
Meanwhile, for the small-object class "Car", the model attains an F1 score of 97.52\%, showcasing robust multi-scale perception and fine-grained segmentation capability in dense urban scenes.

In summary, BiCoR-Seg achieves performance surpassing existing methods across three datasets with distinct characteristics, providing strong evidence of the effectiveness of our proposed approach. This demonstrates that our model can dynamically and precisely inject high-level semantic knowledge into pixel-level feature learning, thereby significantly enhancing segmentation accuracy and generalization capability in complex remote sensing scenarios.

\subsection{Ablation Study}

\paragraph{Ablation on Key Components.}
To systematically validate the effectiveness of the core components in our framework, 
we conduct ablation studies on the HBIS module and the multi-level cooperative supervision losses. We set a model employing a conventional 
\emph{cross-attention} mechanism as the baseline. 
Specifically, we replace the HBIS module in the decoder with a standard cross-attention layer, 
which also enables interactions between class embeddings and pixel feature maps, 
but lacks our heatmap-guided, \emph{explicit bidirectional co-refinement} mechanism. 
The baseline model is optimized only with the main segmentation loss $\mathcal{L}_{main}$.
\begin{table}[t]
\centering
\caption{Ablation study on key components. The checkmark indicates the module or loss is enabled.}
\label{tab:ablation}
\renewcommand{\arraystretch}{1.1}
\setlength{\tabcolsep}{4pt}
\begin{tabular}{cccccc}
\toprule
\multicolumn{4}{c}{\textbf{Modules}} & \multicolumn{1}{c}{\textbf{mIoU}} & \multirow{2}{*}{\textbf{Params (M)}} \\
\cmidrule(lr){1-4} \cmidrule(lr){5-5}
\textbf{Baseline} & \textbf{HBIS} & $\mathcal{L}_{FD}$ & $\mathcal{L}_{HM}$ & \textbf{LoveDA} &  \\
\midrule
\checkmark &  &  &  & 54.20 & 93.5 \\[-2pt]
\cline{6-6}
\checkmark & \checkmark &  &  & 55.15 & \multirow{4}{*}{\centering 89.5} \\
\checkmark &  & \checkmark &  & 55.21 &  \\
\checkmark & \checkmark &  & \checkmark & 55.36 &  \\
\checkmark & \checkmark & \checkmark & \checkmark & \textbf{55.49} &  \\
\bottomrule
\end{tabular}
\end{table}
\begin{table}[t]
\centering
\caption{Ablation on the number of HBIS layers.}
\label{tab:hbis_layers}
\renewcommand{\arraystretch}{1.1}
\setlength{\tabcolsep}{6pt}
\resizebox{\linewidth}{!}{
\begin{tabular}{cccc}
\toprule
\multirow{2}{*}{\textbf{Layers}} & 
\multicolumn{2}{c}{\textbf{mIoU}} & 
\multirow{2}{*}{\textbf{Params(M)}} \\
\cmidrule(lr){2-3}
& \textbf{LoveDA} & \textbf{Vaihingen} & \\
\midrule
$l=1$ & 55.29 & 84.54 & 89.0 \\
$l=2$ & \textbf{55.49} & \textbf{85.37} & 89.5 \\
$l=3$ & 55.28 & 84.67 & 90.0 \\
\bottomrule
\end{tabular}
}
\end{table}

\begin{figure}[t]
    \centering
    \includegraphics[width=\linewidth]{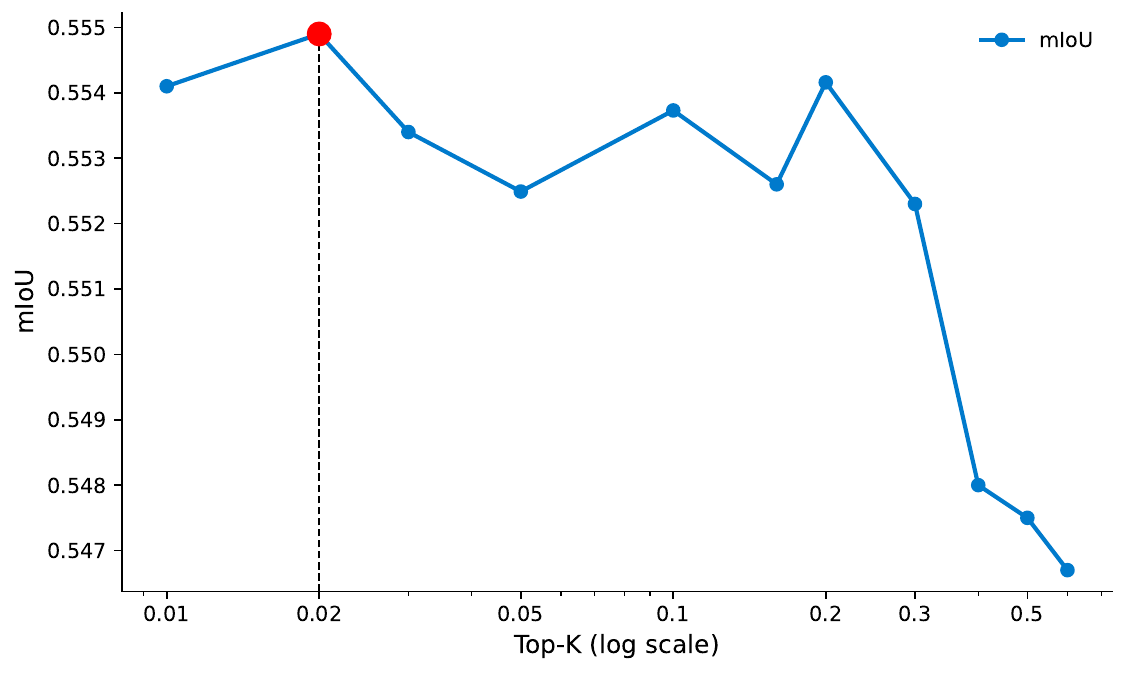}
    \caption{Effect of the Top-K threshold on model performance on the LoveDA dataset.}
    \label{fig:topk}
\end{figure}

\subsection{Quantitative Analysis}
\begin{figure*}[t]
    \centering
    \includegraphics[width=\textwidth]{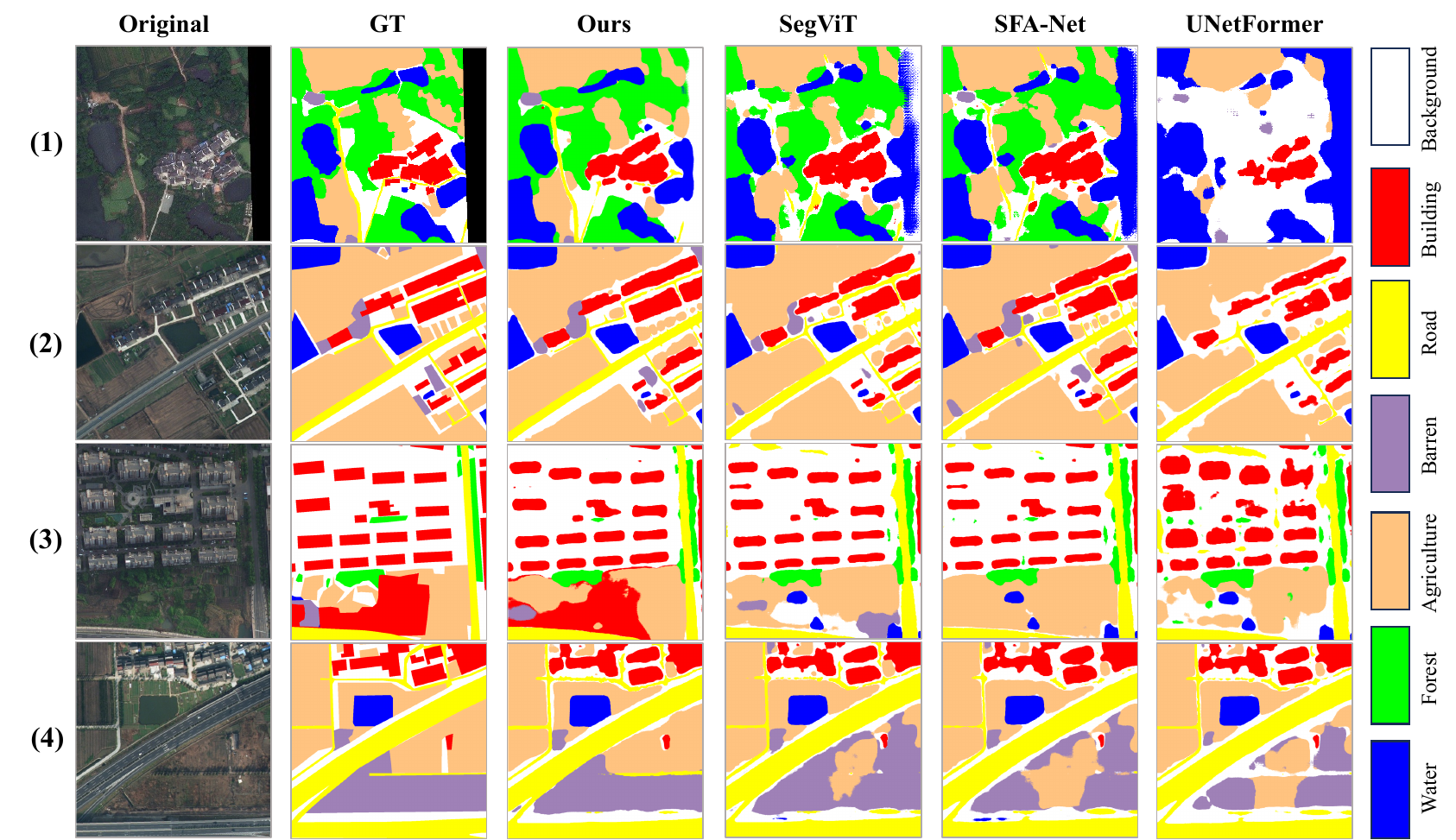}
    \caption{Qualitative comparison of segmentation results on challenging remote sensing scenes. BiCoR-Seg produces more complete and semantically consistent segmentation maps compared with other representative methods. 
    }
    \label{fig:vis_seg}
\end{figure*}
\begin{figure*}[t]
    \centering
    \includegraphics[width=\textwidth]{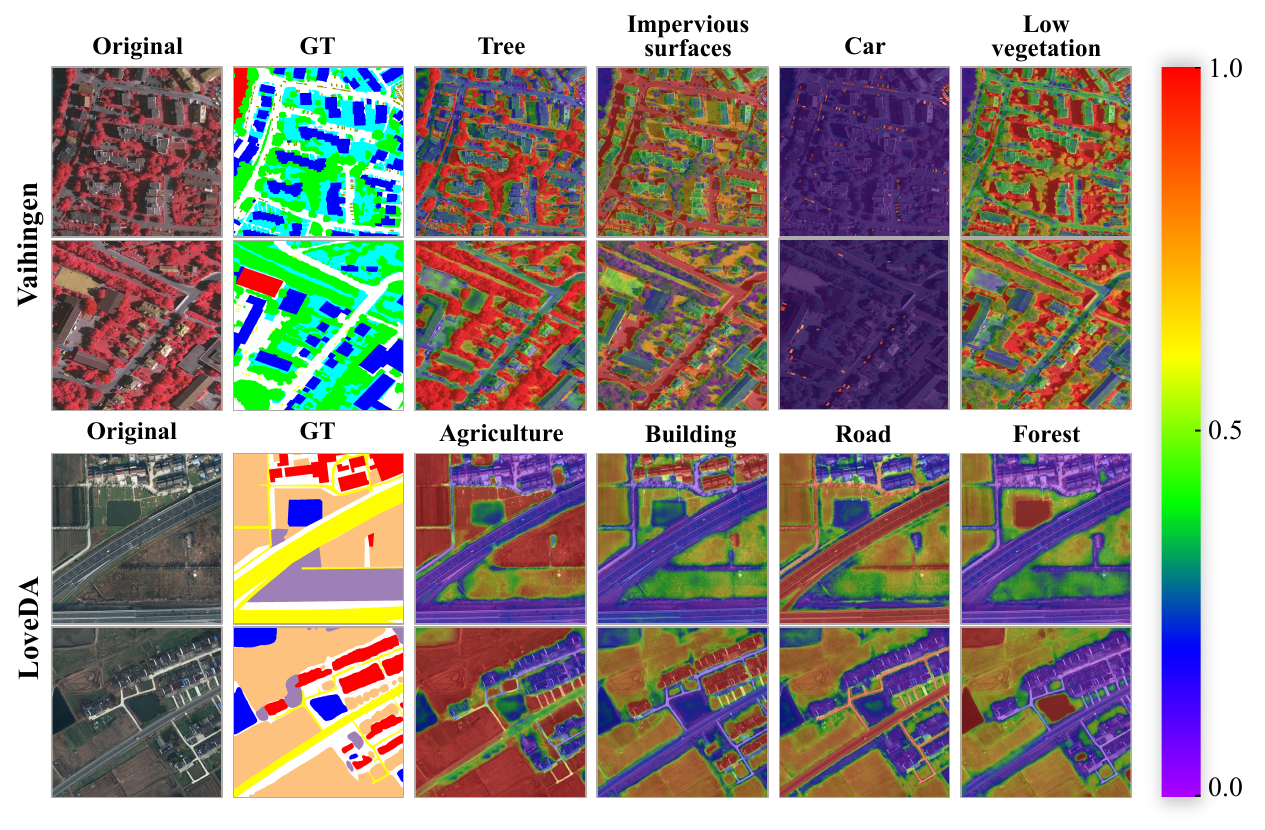}
    \caption{Visualization of class heatmaps generated by BiCoR-Seg for representative categories.
    The HBIS module effectively highlights category-specific regions while suppressing irrelevant background noise.}
    \label{fig:heatmaps}
\end{figure*}
\paragraph{Number of HBIS Modules $L$.}

The number of cascaded HBIS modules $L$ in the decoder determines the depth of semantic--feature interaction and refinement.
We evaluate the model performance with different values of $L$ on the LoveDA and Vaihingen datasets.
As shown in Table~\ref{tab:hbis_layers}, when $L{=}1$, the performance is relatively low, indicating that a single interaction is insufficient for adequate semantic alignment.
When $L$ increases to $2$, the performance improves markedly.
However, when $L$ is further increased to $3$, a slight degradation is observed, which may be attributed to information redundancy caused by overly deep interactions.
Therefore, we set $L{=}2$ in all experiments to strike the best balance between performance and computational efficiency.

As shown in Table~\ref{tab:ablation}, simply replacing the cross-attention module in the baseline 
with our HBIS module yields a notable improvement: 
the mIoU on LoveDA increases from \(\mathbf{54.20\%}\) to \(\mathbf{55.15\%}\). 
This gain strongly demonstrates that, compared with conventional attention-based interaction, 
our bidirectional co-refinement mechanism more effectively establishes 
tight connections between high-level semantics and pixel features. 
We further introduce two auxiliary losses. 
Using either the hierarchical heatmap supervision or the Fisher discriminant loss alone 
consistently improves performance. 
When both are applied together, the model achieves the best results, 
reaching \(\mathbf{55.49\%}\) mIoU on LoveDA. 
These results indicate that the spatial alignment supervision provided by $\mathcal{L}_{HM}$ 
and the semantic discriminability constraint from $\mathcal{L}_{FD}$ are complementary, 
and their combination offers the most comprehensive optimization objective for BiCoR-Seg.

\paragraph{Effect of the Top-K Threshold.}
In the F2CE pathway, we employ a Top-K mechanism to select high-response pixels for updating class embeddings.
The size of the Top-K threshold determines the range of visual feature used for class embedding updates.
A too-small threshold may lead to insufficient information, whereas an excessively large one may introduce noise.
We evaluate the influence of different threshold values on model performance using the LoveDA dataset, and the results are shown in Fig.~\ref{fig:topk}.

Experimental results show that when the Top-K threshold is set to 0.02, 
the model achieves the best performance, reaching an mIoU of 55.49\%.
This indicates that selecting the top 2\% of pixels with the highest responses to update class embeddings 
captures the most representative visual information while effectively filtering out irrelevant regions.

We visualize the segmentation results of BiCoR-Seg and several representative methods on challenging scenes, as shown in Fig.~\ref{fig:vis_seg}.
It can be clearly observed that our method significantly outperforms other segmentation approaches.
For instance, in the fourth example, the comparative methods confuse the purple \textit{bare soil} with the orange \textit{agriculture}, 
while our method generates complete and highly discriminative segmentation masks, demonstrating stronger spatial awareness.

To further illustrate the interpretability of BiCoR-Seg, particularly the working mechanism of the HBIS module, 
we visualize the class heatmaps generated for representative categories, as shown in Fig.~\ref{fig:heatmaps}.
The heatmaps explicitly reveal the spatial regions that the model focuses on during segmentation decision-making.
It can be seen that the heatmaps accurately highlight the pixel regions corresponding to each semantic category 
and effectively suppress background noise.
For the \textit{building} category, the heatmap precisely emphasizes rooftop regions, 
while for small objects such as \textit{vehicles}, the heatmap also achieves accurate localization.
These observations demonstrate that, through iterative collaborative refinement, 
BiCoR-Seg is capable of purifying its internal semantic representations and focusing its attention 
on the most discriminative ground-object regions.
This provides interpretable evidence for its superior segmentation performance, 
visually confirming that our proposed HBIS mechanism enables high-level semantic concepts 
to be progressively and precisely aligned with pixel-level visual evidence, 
thereby significantly improving final segmentation accuracy.

\section{Conclusion}

In this paper, we propose a Bidirectional Collaborative Refinement Framework (BiCoR-Seg) to address the prevalent issues of inter-class similarity, intra-class variability, and semantic ambiguity in high-resolution remote sensing image semantic segmentation.
The framework establishes an explicit semantic association between feature representations and class embeddings through a heatmap-driven bidirectional information collaboration mechanism, enabling the model to achieve progressive optimization of features and semantics across layers.
Experimental results demonstrate that BiCoR-Seg achieves superior performance on multiple benchmark datasets, significantly improving segmentation accuracy and boundary sharpness in complex scenes. Moreover, quantitative analyses show that the class-specific heatmaps accurately reflect the model's discriminative rationale, thereby validating the effectiveness of the bidirectional refinement mechanism.


{\small
\bibliographystyle{cvm}
\bibliography{cvmbib}
}
\end{document}